\pdfoutput=1
\documentclass[11pt]{article}

\usepackage[]{ACL2023}

\usepackage{times}
\usepackage{latexsym}
\usepackage{graphicx}
\usepackage{arydshln}

\usepackage[T1]{fontenc}
\usepackage[utf8]{inputenc}

\usepackage{inconsolata}
\usepackage{array,booktabs,makecell}

\usepackage{ragged2e}
\usepackage{multirow}
\usepackage{graphicx}
\usepackage{color, colortbl}
\usepackage{xstring}
\usepackage{arydshln}

\newcommand{\wojoodfine}{\cal{\textit{Wojood\textsubscript{$Fine$}} }}

\newcommand{\tag}[1]{\texttt{{\small #1}}}

\title{WojoodNER 2024:\\ The Second Arabic Named Entity Recognition Shared Task}

\author{\normalsize \textbf{Mustafa Jarrar}$^\sigma$ ~ \textbf{Nagham Hamad}$^\sigma$~   \textbf{Mohammed Khalilia}$^\sigma$ ~ \textbf{Bashar Talafha}$^\lambda$\\
\normalsize \textbf{AbdelRahim Elmadany}$^\lambda$ ~ \textbf{Muhammad Abdul-Mageed}$^{\lambda,\xi}$ ~ \\
\normalsize $^\sigma$Birzeit University, Palestine\\
  \normalsize $^\lambda$The University of British Columbia\\
  \normalsize  $^\xi$MBZUAI\\ %
  \texttt{\small \{mjarrar,nhamad,mkhalilia\}@birzeit.edu ~  \small  \{btalafha,a.elmadany,muhammad.mageed\}@ubc.ca~ } 
}

\begin{document}
{\makeatletter\acl@finalcopytrue
  \maketitle
}

\begin{abstract}
We present WojoodNER-$2024$, the second Arabic Named Entity Recognition (NER) Shared Task. In WojoodNER-$2024$, we focus on fine-grained Arabic NER. We provided participants with a new Arabic fine-grained NER dataset called \wojoodfine, annotated with subtypes of entities. WojoodNER-$2024$ encompassed three subtasks: ($i$) Closed-Track Flat Fine-Grained NER, ($ii$) Closed-Track Nested Fine-Grained NER, and ($iii$) an Open-Track NER for the Israeli War on Gaza. A total of $43$ unique teams registered for this shared task. Five teams participated in the Flat Fine-Grained Subtask, among which two teams tackled the Nested Fine-Grained Subtask and one team participated in the Open-Track NER Subtask. The winning teams achieved $F_1$ scores of $91\%$ and $92\%$ in the Flat Fine-Grained and Nested Fine-Grained Subtasks, respectively. The sole team in the Open-Track Subtask achieved an $F_1$ score of $73.7\%$.
\end{abstract}

\section{Introduction} 

NER plays a crucial role in various Natural Language Processing (NLP) applications, such as question-answering systems \cite{shaheen2014arabic}, knowledge graphs \cite{james1992knowledge}, and semantic search \cite{guha2003semantic},
information extraction and retrieval \citep{jiang-etal-2016-evaluating}, word sense disambiguation \cite{JMHK23,HJ21b}, machine translation \citep{jain2019entity,nlp23}, automatic summarization \citep{summerscales2011automatic,nlp23},  interoperability \cite{JDF11} and cybersecurity \citep{10.1007/978-3-030-51310-8_2}.

NER involves identifying mentions of named entities in unstructured text and categorizing them into predefined classes, such as \tag{PERS}, \tag{ORG}, \tag{GPE}, \tag{LOC}, \tag{EVENT}, and \tag{DATE}. Given the relative scarcity of resources for Arabic NLP, research in Arabic NER has predominantly concentrated on "flat" entities and has been limited to a few "coarse-grained" entity types, namely \tag{PERS}, \tag{ORG}, and \tag{LOC}. To address this limitation, the WojoodNER shared task series was initiated~\cite{JAKBEHO23}. It aims to enrich Arabic NER research by introducing Wojood and \wojoodfine~\cite{LJKOA23}, nested and fine-grained Arabic NER corpora.
\begin{figure}[t]
  \begin{center}
  % \frame{
  \includegraphics[width=0.95\linewidth]{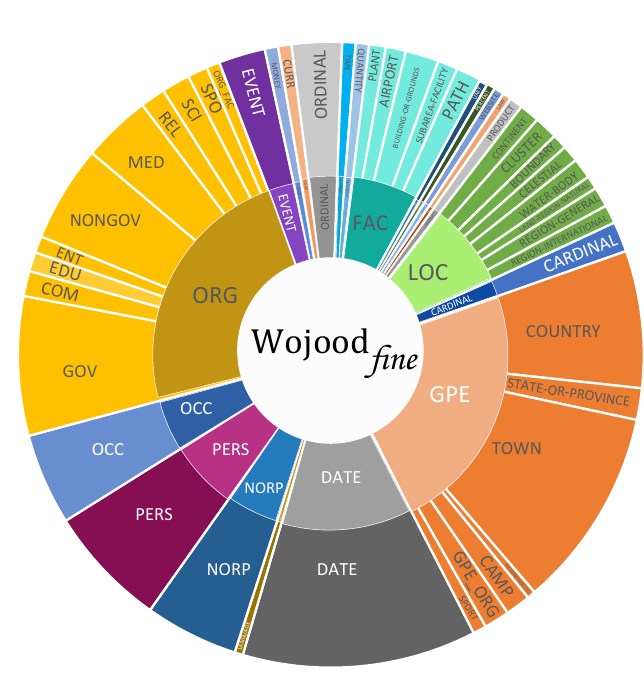}
  % }
  \end{center}
\caption{Visualization of the fine-grained entity types in \wojoodfine}
\label{fig:examples}
\end{figure}

In WojoodNER-$2024$ we provide a new version of Wojood, called \wojoodfine. \wojoodfine enhances the original Wojood corpus by offering fine-grained entity types that are more granular than the data provided in WojoodNER-$2023$. For instance, \tag{GPE} is now divided into seven subtypes: \tag{COUNTRY}, \tag{STATE-OR-PROVINCE}, \tag{TOWN}, \tag{NEIGHBORHOOD}, \tag{CAMP}, \tag{GPE}\_\tag{ORG}, and \tag{SPORT}. \tag{LOC}, \tag{ORG}, and \tag{FAC} are also divided into subtypes as shown in Figure \ref{fig:examples}. \wojoodfine contains approximately $550$k tokens and annotated with $51$ entity types and subtypes, covering $47$k subtype entity mentions. It is worth mentioning that SinaTools supports Wojood and can be accessed via Application Programming Interface (API) \cite{HJK24}.

Teams were invited to experiment with various approaches, ranging from classical machine learning to advanced deep learning and transformer-based techniques, among others. The shared task generated a remarkably diverse array of submissions. A total of $43$ teams registered to participate in the shared task. Among these, five teams successfully submitted their models for evaluation on the blind test set during the final evaluation phase.

The rest of the paper is organized as follows: Section~\ref{sec:literture} provides a brief overview of Arabic NER. We describe the three subtasks and the shared task restrictions in Section~\ref{sec:task_description}. Section~\ref{sec:eval} introduces shared task datasets and evaluation. We present the participating teams, submitted systems and shared task results in Section~\ref{sec:teams_results}. We conclude in Section~\ref{sec:conc}.
\section{Literature Review} \label{sec:literture} 

NER has been an area of active research for many years, witnessing notable progress recently. This section will cover the evolution from initial efforts in recognizing flat-named entities to the current focus on nested NER, with a particular emphasis on Arabic NER, including discussions on corpora, methodologies, and shared tasks. 

\paragraph{Corpora.} The majority of Arabic NER corpora are designed for flat NER annotation. ANERCorp \cite{benajiba2007anersys}, derived from news sources, contains approximately $150k$ tokens and focuses on four specific entity types. CANERCorpus \cite{salah2018building} targets Classical Arabic (CA) and includes a dataset of $258k$ tokens annotated for $14$ types of entities related to religious contexts. The ACE2005 \cite{ACE2005} corpus is multilingual and includes Arabic texts annotated with five distinct entity types. The Ontonotes5 \cite{ontonote} dataset features around $300$k tokens annotated with $18$ different entity types. However, these corpora are somewhat dated and primarily cover media and political domains, which may not accurately reflect contemporary Arabic usage, particularly as language models are sensitive to changes over time and across domains. Recently, \cite{JKG22} introduced Wojood, the largest Arabic NER corpus to date, notable for supporting both flat and nested entity annotations. This corpus, essential for this shared task, includes about $550$k tokens and covers $21$ unique entity types across Modern Standard Arabic (MSA) and two Arabic dialects (Palestinian Curras2 and Lebanese Baladi corpora \cite{EJHZ22}). \wojoodfine \cite{LJKOA23}, an extension of Wojood adds support for entity sub-types, with a total of $51$ entities organized in two-level hierarchy. It is important to note that Wojood has been recently extended to include relationships \cite{JDJK24}. 

\paragraph{Methodologies.} Research in Arabic NER employs a variety of approaches, ranging from rule-based systems \cite{shaalan2007person, jaber2017morphology} to machine learning techniques \cite{settles2004biomedical, abdul-hamid-darwish-2010-simplified, zirikly-diab-2014-named, dahan2015first, DH21}. Recent studies have adopted deep learning strategies, utilizing character and word embeddings in conjunction with Long-Short Term Memory (LSTM) \cite{ali2018bidirectional}, and BiLSTM architectures paired with Conditional Random Field (CRF) layer \cite{el2019arabic, khalifa2019character}. Deep Neural Networks (DNN) are explored in \cite{gridach2018deep}, alongside pretrained Language Models (LM) \cite{JKG22, LJKOA23}. \citet{wang2022nested} conducted a comprehensive review of various approaches to nested entity recognition, including rule-based, layered-based, region-based, hypergraph-based, and transition-based methods. \citet{fei2020dispatched} introduced a multi-task learning framework for nested NER using a dispatched attention mechanism. \citet{ouchi2020instance} developed a method for nested NER that calculates all region representations from the contextual encoding sequence and assigns a category label to each. Readers can also refer to the WojoodNER-$2023$ shared task for DNN techniques used for flat and nested ArabicNER \cite{JAKBEHO23}.

\paragraph{Shared tasks.} While numerous shared tasks exist for NER across different languages and domains, such as MultiCoNER for multilingual complex NER \cite{malmasietal2022semeval} the HIPE-$2022$ for NER and linking in multilingual historical documents \cite{HIPE-2022}, RuNNE-$2022$ for nested NER in Russian \cite{RuNNE-2022}, and NLPCC2022 for entity extraction in the material science domain \cite{NLPCC2022}. WojoodNER-$2023$ for flat and nested Arabic NER \cite{JAKBEHO23}, upon which WojoodNER-$2024$ builds on to offer support for entity sub-types. 

There are several related shared tasks for understanding Arabic MSA and dialects, such as the ArabicNLU for word-sense disambiguation \cite{KMSJAEZ24,JMHK23}, NADI for dialect identification \cite{NADI2023}, AraFinNLP for Cross-dialect Intent detection \cite{AraFinNLP24}, among others.

\section{Task Description } \label{sec:task_description}

WojoodNER-$2024$ confronts the intricacies of Arabic NER with three distinct subtasks: Flat Fine-Grained NER, Nested Fine-Grained NER, and Open-Track NER.  These subtasks provide an evaluation environment, allowing researchers to experiment with diverse approaches for identifying and classifying named entities, along with their subtypes, under controlled (closed) and flexible (open) settings. 

\textbf{Remark}: the Wojood dataset used in WojoodNER-$2023$ \cite{JAKBEHO23} cannot be used in this Shared Task because the two datasets follow different annotation guidelines.

\subsection{Closed-Track Flat Fine-Grained NER}
In this subtask, we provide the \wojoodfine Flat train ($70\%$) and development ($10\%$) datasets. The final evaluation of the submitted contributions from participants is conducted on the test set ($20\%$). The flat NER dataset follows the same split as the nested NER dataset. The key difference in flat NER is that each token is assigned a single tag, corresponding to the first high-level tag assigned in the nested NER dataset, and followed by a single tag in the second level (subtype). This subtask is a closed track, thus participants can only use the provided datasets to train their systems, with no external datasets permitted. 

\subsection{Closed-Track Nested Fine-Grained NER}
This subtask is similar to Subtask 1. We provide the Wojood-Fine Nested train ($70\%$) and development ($10\%$) datasets, with the final evaluation conducted on the test set ($20\%$). This subtask is a closed track, which means participants can only use the provided datasets to train their systems.

\subsection{Open-Track NER - Israeli War on Gaza}
This subtask aims to enable participants to explore the benefits of NER in real-world scenarios. Participants can use external resources and are encouraged to experiment with generative models in various ways, such as fine-tuning, zero-shot learning, and in-context learning. The emphasis on generative models in this subtask is intended to help the Arabic NLP research community gain a better understanding of the capabilities and performance gaps of Large Language Models (LLMs) in information extraction, which is currently a less explored area.

We have curated NER dataset called \textit{Wojood\textsuperscript{Gaza}} pertaining to the ongoing Israeli War on Gaza, based on the assumption that discourse about recent global events will involve mentions from different data distributions. For this subtask, we have collected data from five news domains related to the War, while keeping the identities of these domains confidential. Participants have been provided with a development dataset ($10$k tokens, $2$k from each of the five domains) and a testing dataset ($50$k tokens, $10$k from each domain). Both datasets have been manually annotated with fine-grained named entities, following the same annotation guidelines as in Subtask 1 and Subtask 2, as outlined in \cite{LJKOA23}. This subtask is divided into two subtasks: 3A-flat and 3B-nested.

\subsection{Restrictions}
This section outlines the guidelines for participating in the WojoodNER-$2024$ Shared Task. These rules have been put in place to ensure fairness and transparency for all participants. They also aim to uphold the credibility of the task's assessment process, which is further elaborated on the official shared task FAQ page.

\paragraph{External data.}

For Subtask 1 and 2, participants are strictly forbidden from using external data from previously labeled datasets or employing taggers previously trained to predict named entities. The use of any resources with prior knowledge of NER is not permitted.
On the contrary, Subtask 3 allows the use external resources.

\paragraph{Data format constraints.}
Submissions for the task must be in a single file containing the model's predictions in CoNLL format. This format includes multiple space-separated columns: the first column for tokens and the subsequent columns for tags. For both flat and nested NER, the tag columns follow a predefined order specified on the shared task webpage. The IOB2 scheme \cite{sang1999representing} is used for submissions, consistent with the Wojood dataset. Additionally, text segments are separated by a blank line.

\section{Datasets and Evaluation Metrics} \label{sec:eval} 
In this section, we will describe the dataset, evaluation metrics, and the submission procedure.

\begin{figure}[hbt!]
    \centering
    \includegraphics[width=7.8cm]{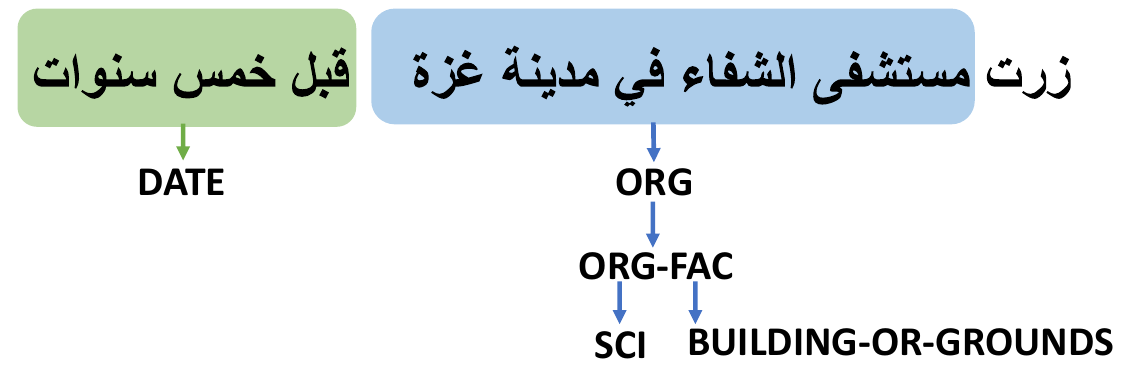}
    \vspace*{-3mm}
    \caption{Flat NER example. }
    \label{fig:flat_ner_example}
\end{figure}

\begin{figure}[hbt!]
    \centering
    \includegraphics[width=7.8cm]{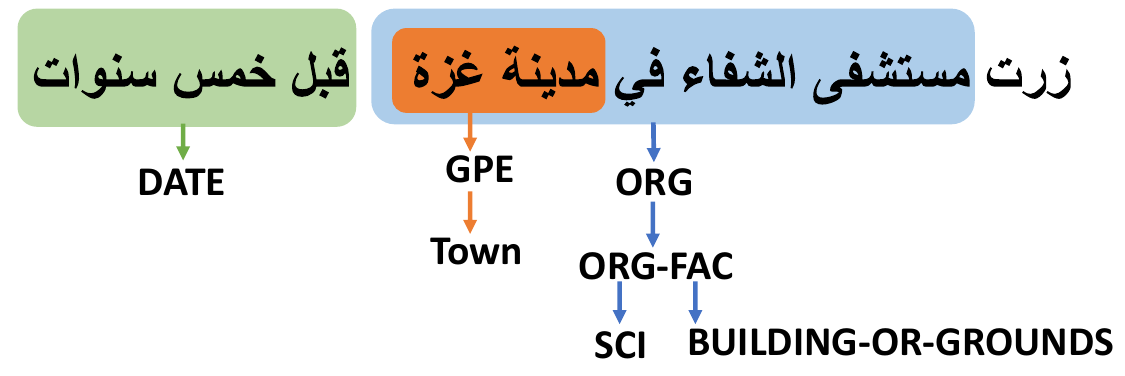}
    \vspace*{-3mm}
    \caption{Nested NER example. }
    \label{fig:nested_ner_example}
\end{figure}

\paragraph{Datasets}
The WojoodNER-$2024$ shared task utilizes the \wojoodfine corpus as a dataset for Subtasks 1 and 2 \cite{LJKOA23}. For Subtask 3, a different dataset called \textit{Wojood\textsuperscript{Gaza}} is utilized that is related to the War on Gaza. The \wojoodfine corpus comprises approximately $550$k tokens, annotated with nest named entities, using $51$ entity types. 
For the purposes of the shared task, we created a flat NER dataset based on the nest NER dataset. That is, the flat NER dataset is created by simplifying the nested NER and reducing subtypes to the top level only as illustrated in Figure \ref{fig:flat_ner_example} and \ref{fig:nested_ner_example}. For both Subtask 1 and Subtask 2, we partitioned the data at the domain level into training, development, and test datasets with a split of $70$:$10$:$20$, respectively.

Table \ref{tab:datasets} presents the details of the datasets used in Subtask 1 (FlatNER) and Subtask 2 (NestNER).


The dataset for Subtask 3 is called \textit{Wojood\textsuperscript{Gaza}}. It includes $60$k tokens that we collected and annotated specifically for this shared task. The corpus was collected from online news articles published at these outlets: 
\href{https://www.palestine-studies.org/ar}{Institute for Palestine Studies}, 
\href{https://www.who.int/ar}{World Health Organization},
\href{https://site.moh.ps/}{Palestinian Ministry of Health},
\href{https://www.pma.ps/ar}{Palestine Monetary Authority},
\href{https://www.aljazeera.net/}{Aljazeera},
\href{https://www.palestineeconomy.ps/ar}{Palestine Economy Portal},
\href
{https://wafa.ps/}{Wafa},
\href{https://www.bnews.ps/ar}{BNews},
\href{https://www.alaraby.com/}{AlAraby},
\href{https://law4palestine.org/ar/}{Law for Palestine},
\href{https://news.un.org/}{United Nations},
\href{https://cnnbusinessarabic.com/}{CNN Business},
\href{https://www.alarabiya.net/}{Al Arabiya},
\href{https://www.skynewsarabia.com/}{Sky News},
\href{https://www.cnbcarabia.com/}{CNBC Arabia},
\href{https://arabic.rt.com/}{RT Arabic}, \href{https://arabic.euronews.com/}{Euro News}, \href{https://www.bbc.com/}{BBC Arabic}.

The articles that were collected from the period of January-March $2024$, covering one of these five domains (politics, law, economy, finance, health) and were directly related to the War on Gaza. For each domain, we collected about $12$k tokens. Participants are provided with the development dataset ($10$k tokens, $2$k from each of the five domains), and a testing dataset ($50$k tokens, $10$k from each domain). Domain names are not provided to the participants.
\textit{Wojood\textsuperscript{Gaza}} was annotated following the same guidelines as \wojoodfine \cite{LJKOA23}.

\begin{table*}[ht!]
\centering
\resizebox{\textwidth}{!}{%
\small
\begin{tabular}{|l|l|rrrr||rrrr|}
\hline
\multicolumn{1}{|c|}{\multirow{2}{*}{Entity Name}} & \multicolumn{1}{c|}{\multirow{2}{*}{NER Tag}} & \multicolumn{4}{c|}{FlatNER} & \multicolumn{4}{c|}{NestedNER} \\ \cline{3-10} 
\multicolumn{1}{|c|}{} & \multicolumn{1}{c|}{} & \multicolumn{1}{c|}{TRAIN} & \multicolumn{1}{c|}{DEV} & \multicolumn{1}{c|}{TEST} & \multicolumn{1}{c|}{Total} & \multicolumn{1}{c|}{TRAIN} & \multicolumn{1}{c|}{DEV} & \multicolumn{1}{c|}{TEST} & \multicolumn{1}{c|}{Total} \\ \hline
Cardinal & \tag{CARDINAL} & \multicolumn{1}{r|}{1291} & \multicolumn{1}{r|}{170} & \multicolumn{1}{r|}{341} & 1802 & \multicolumn{1}{r|}{1312} & \multicolumn{1}{r|}{170} & \multicolumn{1}{r|}{342} & 1824 \\
Organization & \tag{ORG} & \multicolumn{1}{r|}{10590} & \multicolumn{1}{r|}{1488} & \multicolumn{1}{r|}{3006} & 15084 & \multicolumn{1}{r|}{13143} & \multicolumn{1}{r|}{1863} & \multicolumn{1}{r|}{3741} & 18747 \\
Government & \tag{GOV} & \multicolumn{1}{r|}{5689} & \multicolumn{1}{r|}{848} & \multicolumn{1}{r|}{1673} & 8210 & \multicolumn{1}{r|}{5764} & \multicolumn{1}{r|}{860} & \multicolumn{1}{r|}{1695} & 8319 \\
Date & \tag{DATE} & \multicolumn{1}{r|}{10705} & \multicolumn{1}{r|}{1592} & \multicolumn{1}{r|}{3028} & 15325 & \multicolumn{1}{r|}{11346} & \multicolumn{1}{r|}{1691} & \multicolumn{1}{r|}{3206} & 16243 \\
Language & \tag{LANGUAGE} & \multicolumn{1}{r|}{139} & \multicolumn{1}{r|}{16} & \multicolumn{1}{r|}{43} & 198 & \multicolumn{1}{r|}{140} & \multicolumn{1}{r|}{16} & \multicolumn{1}{r|}{43} & 199 \\
Group of people & \tag{NORP} & \multicolumn{1}{r|}{3586} & \multicolumn{1}{r|}{508} & \multicolumn{1}{r|}{1008} & 5102 & \multicolumn{1}{r|}{3952} & \multicolumn{1}{r|}{551} & \multicolumn{1}{r|}{1094} & 5597 \\
Person & \tag{PERS} & \multicolumn{1}{r|}{4519} & \multicolumn{1}{r|}{611} & \multicolumn{1}{r|}{1408} & 6538 & \multicolumn{1}{r|}{5044} & \multicolumn{1}{r|}{677} & \multicolumn{1}{r|}{1565} & 7286 \\
Occupation & \tag{OCC} & \multicolumn{1}{r|}{3717} & \multicolumn{1}{r|}{514} & \multicolumn{1}{r|}{1090} & 5321 & \multicolumn{1}{r|}{3822} & \multicolumn{1}{r|}{532} & \multicolumn{1}{r|}{1124} & 5478 \\
GeoPolitical Entity & \tag{GPE} & \multicolumn{1}{r|}{8052} & \multicolumn{1}{r|}{1116} & \multicolumn{1}{r|}{2395} & 11563 & \multicolumn{1}{r|}{16113} & \multicolumn{1}{r|}{2310} & \multicolumn{1}{r|}{4676} & 23099 \\
Country & \tag{COUNTRY} & \multicolumn{1}{r|}{2911} & \multicolumn{1}{r|}{436} & \multicolumn{1}{r|}{834} & 4181 & \multicolumn{1}{r|}{5744} & \multicolumn{1}{r|}{835} & \multicolumn{1}{r|}{1622} & 8201 \\
Event & \tag{EVENT} & \multicolumn{1}{r|}{1850} & \multicolumn{1}{r|}{282} & \multicolumn{1}{r|}{549} & 2681 & \multicolumn{1}{r|}{1929} & \multicolumn{1}{r|}{292} & \multicolumn{1}{r|}{569} & 2790 \\
Facility & \tag{FAC} & \multicolumn{1}{r|}{560} & \multicolumn{1}{r|}{86} & \multicolumn{1}{r|}{179} & 825 & \multicolumn{1}{r|}{777} & \multicolumn{1}{r|}{116} & \multicolumn{1}{r|}{227} & 1120 \\
Building or ground & \tag{BUILDING-OR-GROUNDS} & \multicolumn{1}{r|}{646} & \multicolumn{1}{r|}{92} & \multicolumn{1}{r|}{193} & 931 & \multicolumn{1}{r|}{706} & \multicolumn{1}{r|}{102} & \multicolumn{1}{r|}{204} & 1012 \\
Town & \tag{TOWN} & \multicolumn{1}{r|}{4970} & \multicolumn{1}{r|}{690} & \multicolumn{1}{r|}{1460} & 7120 & \multicolumn{1}{r|}{8374} & \multicolumn{1}{r|}{1216} & \multicolumn{1}{r|}{2431} & 12021 \\
Loction & \tag{LOC} & \multicolumn{1}{r|}{747} & \multicolumn{1}{r|}{108} & \multicolumn{1}{r|}{234} & 1089 & \multicolumn{1}{r|}{985} & \multicolumn{1}{r|}{141} & \multicolumn{1}{r|}{317} & 1443 \\
Continent & \tag{CONTINENT} & \multicolumn{1}{r|}{65} & \multicolumn{1}{r|}{10} & \multicolumn{1}{r|}{23} & 98 & \multicolumn{1}{r|}{133} & \multicolumn{1}{r|}{23} & \multicolumn{1}{r|}{57} & 213 \\
Money & \tag{MONEY} & \multicolumn{1}{r|}{172} & \multicolumn{1}{r|}{22} & \multicolumn{1}{r|}{33} & 227 & \multicolumn{1}{r|}{172} & \multicolumn{1}{r|}{22} & \multicolumn{1}{r|}{33} & 227 \\
Currency & \tag{CURR} & \multicolumn{1}{r|}{15} & \multicolumn{1}{r|}{2} & \multicolumn{1}{r|}{8} & 25 & \multicolumn{1}{r|}{176} & \multicolumn{1}{r|}{24} & \multicolumn{1}{r|}{41} & 241 \\
Ordinal & \tag{ORDINAL} & \multicolumn{1}{r|}{2739} & \multicolumn{1}{r|}{445} & \multicolumn{1}{r|}{889} & 4073 & \multicolumn{1}{r|}{3444} & \multicolumn{1}{r|}{544} & \multicolumn{1}{r|}{1083} & 5071 \\
Educational & \tag{EDU} & \multicolumn{1}{r|}{440} & \multicolumn{1}{r|}{49} & \multicolumn{1}{r|}{134} & 623 & \multicolumn{1}{r|}{821} & \multicolumn{1}{r|}{109} & \multicolumn{1}{r|}{229} & 1159 \\
Time & \tag{TIME} & \multicolumn{1}{r|}{309} & \multicolumn{1}{r|}{33} & \multicolumn{1}{r|}{84} & 426 & \multicolumn{1}{r|}{311} & \multicolumn{1}{r|}{33} & \multicolumn{1}{r|}{84} & 428 \\
Sports & \tag{SPO} & \multicolumn{1}{r|}{11} & \multicolumn{1}{r|}{2} & \multicolumn{1}{r|}{8} & 21 & \multicolumn{1}{r|}{11} & \multicolumn{1}{r|}{2} & \multicolumn{1}{r|}{8} & 21 \\
Sport & \tag{SPORT} & \multicolumn{1}{r|}{5} & \multicolumn{1}{r|}{2} & \multicolumn{1}{r|}{0} & 7 & \multicolumn{1}{r|}{5} & \multicolumn{1}{r|}{2} & \multicolumn{1}{r|}{1} & 8 \\
Land Region Natural & \tag{LAND-REGION-NATURAL} & \multicolumn{1}{r|}{158} & \multicolumn{1}{r|}{22} & \multicolumn{1}{r|}{52} & 232 & \multicolumn{1}{r|}{179} & \multicolumn{1}{r|}{26} & \multicolumn{1}{r|}{59} & 264 \\
Cluster & \tag{CLUSTER} & \multicolumn{1}{r|}{138} & \multicolumn{1}{r|}{18} & \multicolumn{1}{r|}{55} & 211 & \multicolumn{1}{r|}{222} & \multicolumn{1}{r|}{28} & \multicolumn{1}{r|}{78} & 328 \\
Quantity & \tag{QUANTITY} & \multicolumn{1}{r|}{43} & \multicolumn{1}{r|}{3} & \multicolumn{1}{r|}{9} & 55 & \multicolumn{1}{r|}{46} & \multicolumn{1}{r|}{3} & \multicolumn{1}{r|}{9} & 58 \\
Unit & \tag{UNIT} & \multicolumn{1}{r|}{6} & \multicolumn{1}{r|}{1} & \multicolumn{1}{r|}{2} & 9 & \multicolumn{1}{r|}{46} & \multicolumn{1}{r|}{4} & \multicolumn{1}{r|}{11} & 61 \\
State-or-Province & \tag{STATE-OR-PROVINCE} & \multicolumn{1}{r|}{1146} & \multicolumn{1}{r|}{159} & \multicolumn{1}{r|}{372} & 1677 & \multicolumn{1}{r|}{1292} & \multicolumn{1}{r|}{179} & \multicolumn{1}{r|}{421} & 1892 \\
Non-Governmental & \tag{NONGOV} & \multicolumn{1}{r|}{4030} & \multicolumn{1}{r|}{566} & \multicolumn{1}{r|}{1143} & 5739 & \multicolumn{1}{r|}{4071} & \multicolumn{1}{r|}{573} & \multicolumn{1}{r|}{1158} & 5802 \\
Neighborhood & \tag{NEIGHBORHOOD} & \multicolumn{1}{r|}{78} & \multicolumn{1}{r|}{5} & \multicolumn{1}{r|}{29} & 112 & \multicolumn{1}{r|}{87} & \multicolumn{1}{r|}{5} & \multicolumn{1}{r|}{30} & 122 \\
Water-Body &\tag{WATER-BODY} & \multicolumn{1}{r|}{76} & \multicolumn{1}{r|}{14} & \multicolumn{1}{r|}{18} & 108 & \multicolumn{1}{r|}{88} & \multicolumn{1}{r|}{14} & \multicolumn{1}{r|}{21} & 123 \\
Percent & \tag{PERCENT} & \multicolumn{1}{r|}{92} & \multicolumn{1}{r|}{12} & \multicolumn{1}{r|}{33} & 137 & \multicolumn{1}{r|}{92} & \multicolumn{1}{r|}{12} & \multicolumn{1}{r|}{33} & 137 \\
Camp & \tag{CAMP} & \multicolumn{1}{r|}{595} & \multicolumn{1}{r|}{69} & \multicolumn{1}{r|}{167} & 831 & \multicolumn{1}{r|}{605} & \multicolumn{1}{r|}{71} & \multicolumn{1}{r|}{168} & 844 \\
Path & \tag{PATH} & \multicolumn{1}{r|}{52} & \multicolumn{1}{r|}{6} & \multicolumn{1}{r|}{18} & 76 & \multicolumn{1}{r|}{52} & \multicolumn{1}{r|}{6} & \multicolumn{1}{r|}{18} & 76 \\
Media & \tag{MED} & \multicolumn{1}{r|}{2886} & \multicolumn{1}{r|}{419} & \multicolumn{1}{r|}{807} & 4112 & \multicolumn{1}{r|}{2886} & \multicolumn{1}{r|}{419} & \multicolumn{1}{r|}{807} & 4112 \\
Region-General & \tag{REGION-GENERAL} & \multicolumn{1}{r|}{275} & \multicolumn{1}{r|}{37} & \multicolumn{1}{r|}{67} & 379 & \multicolumn{1}{r|}{278} & \multicolumn{1}{r|}{37} & \multicolumn{1}{r|}{69} & 384 \\
GPE\_ORG & \tag{GPE\_ORG} & \multicolumn{1}{r|}{1000} & \multicolumn{1}{r|}{161} & \multicolumn{1}{r|}{316} & 1477 & \multicolumn{1}{r|}{1036} & \multicolumn{1}{r|}{167} & \multicolumn{1}{r|}{325} & 1528 \\
Website & \tag{WEBSITE} & \multicolumn{1}{r|}{412} & \multicolumn{1}{r|}{80} & \multicolumn{1}{r|}{116} & 608 & \multicolumn{1}{r|}{412} & \multicolumn{1}{r|}{80} & \multicolumn{1}{r|}{116} & 608 \\
Commercial & \tag{COM} & \multicolumn{1}{r|}{458} & \multicolumn{1}{r|}{39} & \multicolumn{1}{r|}{111} & 608 & \multicolumn{1}{r|}{459} & \multicolumn{1}{r|}{40} & \multicolumn{1}{r|}{111} & 610 \\
Celectial & \tag{CELESTIAL} & \multicolumn{1}{r|}{2} & \multicolumn{1}{r|}{0} & \multicolumn{1}{r|}{2} & 4 & \multicolumn{1}{r|}{2} & \multicolumn{1}{r|}{0} & \multicolumn{1}{r|}{2} & 4 \\
Subarea - Facility & \tag{SUBAREA-FACILITY} & \multicolumn{1}{r|}{91} & \multicolumn{1}{r|}{16} & \multicolumn{1}{r|}{23} & 130 & \multicolumn{1}{r|}{96} & \multicolumn{1}{r|}{16} & \multicolumn{1}{r|}{23} & 135 \\
Medical-Science & \tag{SCI} & \multicolumn{1}{r|}{102} & \multicolumn{1}{r|}{12} & \multicolumn{1}{r|}{29} & 143 & \multicolumn{1}{r|}{104} & \multicolumn{1}{r|}{13} & \multicolumn{1}{r|}{30} & 147 \\
Religious & \tag{REL} & \multicolumn{1}{r|}{61} & \multicolumn{1}{r|}{10} & \multicolumn{1}{r|}{24} & 95 & \multicolumn{1}{r|}{61} & \multicolumn{1}{r|}{10} & \multicolumn{1}{r|}{25} & 96 \\
ORG\_FAC & \tag{ORG\_FAC} & \multicolumn{1}{r|}{87} & \multicolumn{1}{r|}{7} & \multicolumn{1}{r|}{19} & 113 & \multicolumn{1}{r|}{87} & \multicolumn{1}{r|}{7} & \multicolumn{1}{r|}{19} & 113 \\
Region-International & \tag{REGION-INTERNATIONAL} & \multicolumn{1}{r|}{67} & \multicolumn{1}{r|}{12} & \multicolumn{1}{r|}{29} & 108 & \multicolumn{1}{r|}{70} & \multicolumn{1}{r|}{12} & \multicolumn{1}{r|}{29} & 111 \\
Entertainment & \tag{ENT} & \multicolumn{1}{r|}{1} & \multicolumn{1}{r|}{1} & \multicolumn{1}{r|}{1} & 3 & \multicolumn{1}{r|}{1} & \multicolumn{1}{r|}{1} & \multicolumn{1}{r|}{1} & 3 \\
Boundary & \tag{BOUNDARY} & \multicolumn{1}{r|}{15} & \multicolumn{1}{r|}{4} & \multicolumn{1}{r|}{3} & 22 & \multicolumn{1}{r|}{15} & \multicolumn{1}{r|}{4} & \multicolumn{1}{r|}{3} & 22 \\
Plant & \tag{PLANT} & \multicolumn{1}{r|}{1} & \multicolumn{1}{r|}{0} & \multicolumn{1}{r|}{0} & 1 & \multicolumn{1}{r|}{1} & \multicolumn{1}{r|}{0} & \multicolumn{1}{r|}{0} & 1 \\
Law & \tag{LAW} & \multicolumn{1}{r|}{368} & \multicolumn{1}{r|}{47} & \multicolumn{1}{r|}{90} & 505 & \multicolumn{1}{r|}{368} & \multicolumn{1}{r|}{47} & \multicolumn{1}{r|}{90} & 505 \\
Product & \tag{PRODUCT} & \multicolumn{1}{r|}{61} & \multicolumn{1}{r|}{8} & \multicolumn{1}{r|}{17} & 86 & \multicolumn{1}{r|}{62} & \multicolumn{1}{r|}{8} & \multicolumn{1}{r|}{19} & 89 \\
Airport & \tag{AIRPORT} & \multicolumn{1}{r|}{5} & \multicolumn{1}{r|}{0} & \multicolumn{1}{r|}{1} & 6 & \multicolumn{1}{r|}{5} & \multicolumn{1}{r|}{0} & \multicolumn{1}{r|}{1} & 6 \\ 
 \hline
 & Total & \multicolumn{1}{r|}{76034} & \multicolumn{1}{r|}{10850} & \multicolumn{1}{r|}{22173} & 109057 & \multicolumn{1}{r|}{96947} & \multicolumn{1}{r|}{13913} & \multicolumn{1}{r|}{28068} & 138928 \\
\bottomrule
\end{tabular}% 
}
\caption{ Distribution of NER tags in WojoodNER-2024 Subtask1 (i.e., FlatNER) and Subtask2 (i.e., NestedNER)
across the training (i.e., TRAIN) , development (i.e., DEV), and test (i.e., TEST) splits for the WojoodNER-2024.}\label{tab:datasets}

\end{table*}

\paragraph{Evaluation metrics.} 
The official and primary evaluation metric for Subtask 1, Subtask 2, and Subtask 3 is the micro-averaged $F_1$ score. In addition to this metric, we also report system performance in terms of Precision, Recall, and Accuracy.

\paragraph{Submission rules.} 

Participating teams were allowed to submit up to four runs for each test set across the three subtasks. For each team's submissions, we retained only the highest score per task. Although the official results were derived exclusively from the blind test set, we streamlined the evaluation process by establishing four separate CodaLab competitions, one for each subtask\footnote{The different CodaLab competitions are available at the following links: \href{https://codalab.lisn.upsaclay.fr/competitions/18358}{\texttt{Subtask 1}}, \href{https://codalab.lisn.upsaclay.fr/competitions/11750}{\texttt{Subtask 2}} and
\href{https://codalab.lisn.upsaclay.fr/competitions/18374}{\texttt{Subtask 3A},
\href{https://codalab.lisn.upsaclay.fr/competitions/18384}{\texttt{Subtask 3B}.
}
}}.
 We are keeping the CodaLab for each subtask active even after the official competition has concluded. This is aimed at facilitating researchers who wish to continue training models and evaluating systems with the shared task's blind test sets. As a result, we will not disclose the ground truth labels for the test sets for any of the subtasks.

\section{Shared Task Teams \& Results}\label{sec:teams_results}
\subsection{Participating Teams}

Overall, we received $43$ unique team registrations, $26$ of them registered in the CodaLab, and only seven teams have submitted their results. These seven teams have submitted $263$ valid entries during the testing phase.  Specifically, $76$ submissions for FlatNER were received from six teams, $168$ submissions for NestedNER came from four teams, eight submissions for Gaza-Flat from one team, and $11$ submissions for Gaza-Nested from $1$ team. Table~\ref{tab:teams} provides details about the teams, their affiliations, and their tasks (1-- FlatNER, 2-- NestedNER, 3A-- Gaza-Flat, and 3B-- Gaza-Nested). Out of the seven teams, we received six description papers, which are all accepted for publication.

\begin{table*}[]
\centering

\resizebox{\textwidth}{!}{%
\begin{tabular}{lll}
\hline
\textbf{Team}                               & \textbf{Affiliation(s)}                                                              & \textbf{Task} \\ \hline
\textbf{\texttt{Addax}}  \cite{Addax} & Um6p College Of Computing, Morocco & 1\\

\textbf{\texttt{Bangor University}}  \cite{Bangor} & Bangor University, UK & 1           \\
\textbf{\texttt{DRU}}   \cite{DRU,DRU2} & Arab Center for Research and Policy Studies, Qatar & 1,2,3         \\
\textbf{\texttt{mucAI}} \cite{mucAI}             & \begin{tabular}[c]{@{}l@{}}Technical University of Munich, Germany\\ Helwan University of Cairo, Egypt\end{tabular} & 1         \\
\textbf{\texttt{muNERa}}  \cite{muNERa} & 

\begin{tabular}[c]{@{}l@{}}King Abdulaziz City for Science and Technology (KACST),\\ Saudi Data and Artificial Intelligence Authority (SDAIA),\\ and King Salman Global Academy for Arabic Language (KSGAAL), Saudi Arabia \end{tabular}
& 1,2         \\

\hline
\end{tabular}
}
\caption{List of teams that participated in the WojoodNER-$2024$ subtasks.} 
\label{tab:teams}

\end{table*}

\subsection{Baselines}

For Subtask 1 and Subtask 2, we fine-tuned the AraBERT\textsubscript{v2} \cite{antoun2020arabert} pre-trained model using subtask-specific training data for 20 epochs, with a learning rate of $1e^{-5}$ and a batch size of $8$. To ensure optimal model performance, we incorporated early stopping with a patience setting of $5$. After each epoch, we evaluated the model's performance and selected the best-performing checkpoints based on their performance on the respective development sets. We then present the performance metrics of the best-performing model on the test datasets.

\subsection{Results}
Table~\ref{tab:results1}, Table~\ref{tab:results2}, and Table~\ref{tab:results3} presents the leaderboards for Subtask 1--FlatNER, Subtask 2--NestedNER, and Subtask 3A--Gaza respectively, organized in descending order based on the micro-$F_1$ scores. The micro-$F_1$ score listed for each team reflects their highest score out of the four allowed submissions for each task.
% \textcolor{red}{What about Gaza-subtask3-A?}

\begin{table}[h]
\centering
\resizebox{0.45\textwidth}{!}{%
\begin{tabular}{@{}clrrr@{}}
\hline
\textbf{Rank} & \textbf{Team}   & \textbf{$F_1$} & \textbf{Pre.} & \textbf{Rec.} \\ 
\hline
1 &  mucAI      & $91$ & $91$ & $90$  \\
2 &  muNERa       & $90$ & $91$ & $89$  \\
2 &  Addax       & $90$ & $89$ & $91$  \\
\cdashline{1-5}
& Baseline-I (ARBERT\textsubscript{v2} ) & $89$ & $89$ & $90$  \\
\cdashline{1-5}
3 & DRU - Arab Center & $87$ & $86$ & $86$  \\
4 & Bangor & $86$ & $88$ & $85$  \\
\hline
\end{tabular}%
}
\caption{Results of Subtask 1--FlatNER. 
}
\label{tab:results1}
\end{table}
\textbf{For FlatNER,} the \texttt{mucAI} team~\cite{mucAI} achieved the highest $F_1$ score of $91$, with \texttt{muNERa} \cite{muNERa} and \texttt{Addax} \cite{Addax} securing second place with $90$, \texttt{DRU} taking third place with $87$, and \texttt{Bangor} taking fourth place with $86$. Notably, three teams outperformed our baseline, as shown in Table \ref{tab:results1}. 
The winning team \texttt{mucAI}\cite{mucAI} surpassed the baseline by $2$\%. The performance gap between our baseline and the lowest-performing model is approximately $3$\%. Furthermore, the difference in $F_1$ scores among the teams is minimal, with a standard deviation of $\sigma=1.94$.

\begin{table}[h]
\centering
\resizebox{0.45\textwidth}{!}{%
\begin{tabular}{@{}clrrr@{}}
\hline
\textbf{Rank} & \textbf{Team}   & \textbf{F1} & \textbf{Pre.} & \textbf{Rec.} \\ 
\hline
& Baseline-I (ARBERT\textsubscript{v2} ) & $92$ & $92$ & $93$  \\
\cdashline{1-5}
2 &  muNERa       & $91$ & $92$ & $90$  \\
3 & DRU - Arab Center & $90$ & $90$ & $90$  \\
\hline
\end{tabular}%
}
\caption{Results of Subtask 2 -- NestedNER. 
}
\label{tab:results2}
\end{table}

\textbf{For NestedNER}, none of the teams outperformed the baseline. The \texttt{muNERa} team~\cite{muNERa} achieved the highest $F_1$ score of $91$, but still $1$\% below the baseline, followed by \texttt{DRU} team~\cite{DRU} with a score of $90$.

\begin{table}[h]
\centering
\resizebox{0.45\textwidth}{!}{%
\begin{tabular}{@{}clrrr@{}}
\hline
\textbf{Rank} & \textbf{Team}   & \textbf{$F_1$} & \textbf{Pre.} & \textbf{Rec.} \\ 
\hline
1 &  DRU - Arab Center       & $73.7$ & $71.9$ & $75.6$  \\
\hline
\end{tabular}%
}
\caption{Results of Subtask 3 -- Gaza-FlatNER. 
}
\label{tab:results3}
\end{table}

\textbf{For the open-track Gaza-FlatNER,} only \texttt{DRU} team \cite{DRU} reported their results with a recall of $75.9$ and $F_1$ score of $73.5$. 

\subsection{General Description of Submitted Systems}
For Subtask 1 and Subtask 2, all models submitted to the shared task employed the transfer learning approach, utilizing pre-trained models trained on diverse data sources. For Subtask 3, LLMs with in-context learning techniques were utilized.

% \textcolor{red}{Do we need to write the summary table?!}

\texttt{Addax}~\cite{Addax} proposed a combined tagging approach that merges the main entity type and its subtypes into a single category (e.g., "\texttt{B-GPE+B-COUNTRY}" for "Palestine"). This method follows the IOB2 scheme for entity boundaries and simplifies training by focusing on a single combined tag per entity, integrating both main and subtype information. The model architecture utilizes a two-channel parallel hybrid neural network with an attention mechanism. It employs BERT-based model (AraBERTv0.2-Twitter) embeddings for contextualized word representations and consists of two distinct channels: one using Conv1D layers for local feature extraction and another with Bi-GRU layers to capture long-range dependencies. Additionally, an attention layer focusing on the most relevant input features has been added in each channel. 

\texttt{Bangor}~\cite{Bangor} added a linear layer on top of a BERT-based model (bert-base-arabic-camelbert-mix) to classify each token into one of $51$ different entity types and subtypes, as well as the "\texttt{O}" label for non-entity tokens. This linear layer maps the contextualized embeddings produced by BERT to the desired output labels.

\texttt{muNERa}~\cite{muNERa} team adapted Wojood dataset to fit the input requirements of the Translation between Augmented Natural Languages (TANL) framework \cite{paolini2021structured}. The preprocessing steps included extracting hierarchical tags (parent, subtype, sub-subtype) and their spans using the IOB2 scheme. Each token and its corresponding labels were reformatted to align with the TANL framework's specifications. TANL was used for Subtask 1 and Subtask 2. In this framework, both input and output are structured in augmented natural languages and enclosed in square brackets (e.g., [ token | entity type ]). For nested entities, TANL can represent entity hierarchies, such as [ token [ token | entity type1 ] | entity type2 ]. They utilized two distinct TANL models for handling flat and nested entities. A decoder-encoder model (AraT5v2) is used as base for the TANL model. Additionally, they used a FastText (FT) classifier as a secondary tagger, first using TANL to detect spans and assign level-1 (parent) tags, and then applying the FT classifier to tag the detected spans with level-2 and level-3 tags. The best-performing TANL architecture was achieved without using FT. 

\texttt{mucAI}~\cite{mucAI} team proposed a two-step methodology: joint vanilla fine-tuning followed by $k$-Neared Neighbor (KNN) at inference time. BERT (AraBERTv02) is used as the backbone for generating word embeddings. These embeddings are then fed into two multi-layer perceptrons (MLP) that are trained jointly. The first head predicts one of the predefined $21$ main entity tags. The second head predicts one of the predefined $31$ sub-entities. A “Datastore” is constructed as a database that has a contextualized representation for each token alongside the label in each sentence in the training set. The "Datastore" was queried during inference to retrieve the $k$ nearest neighbors based on a similarity score, derive the distribution of labels from these neighbors, and then interpolate this distribution with the main MLP model's distribution using an interpolation factor to obtain the final label probabilities.

\texttt{DRU-Arab Center}~\cite{DRU}  proposed three strategies to deal with the Flat and Nested subtasks. ($1$) A single-layer approach, where they fine-tuned different BERT-based models to predict all types and subtypes in one shot, using a $103$-length one-hot encoded vector for each type and subtype, including the "\texttt{O}" tag. They experimented with GEMMA \cite{team2024gemma}, and AraBERTv2 \cite{antoun2020arabert}, and fine-tuned BLOOMZ-7b-mt on a high-quality Arabic dataset  \cite{muennighoff-etal-2023-crosslingual}. ($2$) Another attempt was the One×1 classifier method, which separated type and subtype classification by dedicating a model for each, training one instance of (AraBERTv2) exclusively for predicting main types and another instance for predicting sub-types. ($3$) In the One×4 Classifier Method, instead of only one model for subtypes, they trained four instances, each specialized in the sub-types of a specific group: \tag{GPE}, \tag{ORG}, \tag{FAC}, \tag{LOC}, as the other main types have no subtypes. Among these strategies, the One×1 approach achieved the highest performance on both Subtask 1 and Subtask 2.

For the open track Subtask  3, \cite{DRU2}, \texttt{DRU-Arab Center}  utilized LLMs (\textit{Cohere’s Command R model} \cite{cohere_command_r}) and in-context learning to solve this task. In the prompt design, they wrote a detailed system prompt that outlines the steps for tagging tokens according to the \wojoodfine annotation guidelines. The prompt instructs the LLM to perform NER for Arabic text by predicting up to three levels of tags—high-level tags, subtypes, and specific subtypes for certain entities—while simplifying the task to two tag levels for practical purposes, and outputting predictions in CSV format; illustrative examples are provided to guide the model, and specific instructions ensure the correct application of the IOB2 schema and handle complex subtypes during post-processing. Command R's output quality issues included producing extra or missing tokens. To solve that, they post-processed the generated output to match the expected format by assigning the tag "\texttt{O}" to ground truth tokens without corresponding predicted tokens or hallucinated tags, and by converting the remaining format issues to the expected output.

\section{Conclusion}\label{sec:conc}

In this paper, we present the outcomes of the second edition of WojoodNER shared task. The results from the participating teams highlight the ongoing difficulties in NER, yet it is encouraging to see that various innovative approaches, particularly those leveraging the power of language models, have proven effective in tackling this complex task. As we progress, we are dedicated to advancing research in this field. Our vision includes continuous efforts to improve Arabic NER, drawing on the valuable insights from WojoodNER-$2024$ and exploring new solutions. Additionally, we plan to expand the \wojoodfine corpus to encompass more dialects.

\section*{Limitations}\label{sec:limits} 

Similar to WojoodNER-2023, WojoodNER-2024 aimed for the broadest possible coverage, primarily focusing on MSA data. This dataset used this year, \wojoodfine, includes limited data from dialects. It only includes text from Palestinian and Lebanese Arabic. We plan to include the other dialects, especially the Syrian \textit{Nabra} dialects \cite{ANMFTM23} as well as the four dialects in the \textit{Lisan} \cite{JZHNW23} corpus. Additionally, the \textit{Wojood\textsuperscript{Gaza}} dataset used in Subtask 3 covers only the initial phase of the Israeli War on Gaza, excluding the subsequent genocidal and starvation events.

\section*{Ethics Statement}\label{sec:Ethics} 

The datasets provided for this shared task are derived from public sources, eliminating specific privacy concerns. The results of the shared task will be made publicly available to enable the research community to build upon them for the public good and peaceful purposes. Our datasets and research are strictly intended for non-malicious, peaceful, and non-military purposes.

\section*{Acknowledgements}
This research is partially funded by the Palestinian Higher Council for Innovation and Excellence and by the research committee at Birzeit University. 

Muhammad Abdul-Mageed acknowledges support from Canada Research Chairs (CRC), the Natural Sciences and Engineering Research Council of Canada (NSERC; RGPIN-2018-04267), the Social Sciences and Humanities Research Council of Canada (SSHRC; 435-2018-0576; 895-2020-1004; 895-2021-1008), Canadian Foundation for Innovation (CFI; 37771), Digital Research Alliance of Canada,\footnote{\href{https://alliancecan.ca}{https://alliancecan.ca}} and UBC ARC-Sockeye.

We extend our gratitude to Taymaa Hammouda for the technical support and to the students who helped and supported us during the annotation process, especially Haneen Liqreina, Lina Duaibes, Shimaa Hamayel, Rwaa Assi, Hiba Zayed, and Sana Ghanim.

\bibliography{MyReferences,custom}
\bibliographystyle{acl_natbib}

\end{document}